\pdfoutput=1

\documentclass[11pt]{article}

\usepackage{acl}

\usepackage{times}
\usepackage{lipsum}
\usepackage{latexsym}
\usepackage{subcaption}
\usepackage{graphicx}
\usepackage{multirow}
\usepackage{tabularx}
\usepackage{enumitem}
\usepackage{array}
\usepackage{amsmath}
\usepackage[export]{adjustbox}

\usepackage[T1]{fontenc}

\usepackage[utf8]{inputenc}
\usepackage{float}

\usepackage{microtype}
\usepackage{graphicx}
\usepackage{url}
\usepackage{listings}

\usepackage[ruled,longend]{algorithm2e}
\usepackage{amssymb}

\usepackage{xcolor}

\newcommand{\teach}{\textit{TEACh }}
\newcommand{\commander}{\texttt{Commander} }
\newcommand{\follower}{\texttt{Follower} }

\title{\textsc{Interaction is all You Need?}\\A Study of Robots Ability to Understand and Execute}

\author{Kushal Koshti \and Nidhir Bhavsar \\
        Universität Potsdam \\
        \texttt{\{kushal.koshti, nidhir.bhavsar\}@uni-potsdam.de}}     

\begin{document}
\maketitle
\begin{abstract}
This paper aims to address a critical challenge in robotics, which is enabling them to operate seamlessly in human environments through natural language interactions. Our primary focus is to equip robots with the ability to understand and execute complex instructions in coherent dialogs to facilitate intricate task-solving scenarios. To explore this, we build upon the Execution from Dialog History (EDH) task from the \teach benchmark. We employ a multi-transformer model with BART LM. We observe that our best configuration outperforms the baseline with a success rate score of $8.85$ and a goal-conditioned success rate score of $14.02$. In addition, we suggest an alternative methodology for completing this task. Moreover, we introduce a new task by expanding the EDH task and making predictions about game plans instead of individual actions. We have evaluated multiple BART models and an LLaMA2 LLM, which has achieved a ROGUE-L score of $46.77$ for this task. We make our findings accessible over here\footnote{\url{https://github.com/Nid989/TEACh_EDH}}.
\end{abstract}

\section{Introduction}

To enable robots to operate efficiently in human environments, they need the capability to engage in natural language interactions. Achieving this has long been a goal in modeling AI agents, focusing on seamless interaction with humans to assist in task-solving. This involves understanding and executing instructions and facilitating dialog to clarify doubts and correct mistakes \cite{skrynnik2022learning}.

In general, robots operating in the field should acquire the ability to synchronize natural language with their operational environment, which inherently ties into the concept of symbol grounding, concentrating on establishing connections between language and static images \cite{HARNAD1990335}. Over the years, various proposals have emerged to address the challenge of human-AI collaboration, primarily focusing on humans instructing agents to achieve specific goals \citep{DBLP:journals/corr/abs-1912-01734, 10.7551/mitpress/11956.001.0001}. Typically, these tasks are executed in a simulated environment such as the Minecraft virtual environment \cite{narayan-chen-etal-2019-collaborative}, AI2-THOR \cite{DBLP:journals/corr/abs-1712-05474}, AI Habitat \cite{DBLP:journals/corr/abs-1904-01201}, or Matterport 3D \cite{DBLP:journals/corr/abs-1711-07280}. These environments confine the agent within limited boundaries, thus accelerating its learning capabilities.

The introduction of \textit{TEACh} (\textbf{T}ask-driven \textbf{E}mbodied \textbf{A}gents that \textbf{C}hat) by \citet{DBLP:journals/corr/abs-2110-00534} has brought forth benchmarks that effectively address the aforementioned challenges, paving the way for the development of conversational robotics in household tasks.

In this paper, we aim to address the strategy of using trained models to attain elements of embodied intelligence. Similar to the \teach benchmark dataset, we tackle the task of  Execution from Dialog history (EDH). This involves utilizing the concept of instruction-oriented multi-party dialog conversation and employing the ego-centric camera observations perceived by the robot agent to carry out the desired task.  Furthermore, we analyze the dataset to uncover nuances and elements that hinder the performance of the modeled system. 

Overall, the problem aims at agents needing to understand the task instructions provided at varying levels of abstraction across dialogs. We conduct experiments aimed at achieving the task of EDH through the utilization of language generation techniques. Consequently, we provide an explanation regarding the potential unsuitability of this approach for the given dataset. Finally, We try to gain valuable insights by leveraging the natural language component of the dataset to assist in creating human-understandable plans for executing a given task. 

To summarize, our main contributions are:
\begin{itemize}
    \item \textbf{C1}: Modifying the baseline EDH (Execution from Dialog History) implementation with GPLM-based language encodings and adjusted multimodal fusion.
    \item \textbf{C2}: Introducing an alternative approach for the task and proposing a modeling technique for it.
    \item \textbf{C3}: Implementing a language-orientated task to generate game plans constituting action-object pairs based on dialog instructions.
\end{itemize}

\section{Related Work}

\subsection{Vision \& Dialog Navigation and Task Completion}

Certain embodied agent tasks demand the incorporation of visual observations and language instructions to model behavior, particularly in executing navigational actions such as turning or moving forward. The development of models for such behavior within controlled environments has progressed significantly, transitioning from using symbolic environments \citep{inproceedingsmac, 10.5555/2900423.2900560} to adopting photorealistic indoor \cite{DBLP:journals/corr/abs-1711-07280} and outdoor \cite{narayan-chen-etal-2019-collaborative} settings.

Accordingly, some tasks entail combining such navigational actions to accomplish a specific goal. These models encompass a combination of rules and learned components that enable them to understand language \cite{DBLP:journals/corr/abs-1712-01097}. Others operate within a fully observable blocks world \citep{DBLP:journals/corr/abs-1712-03463, misra-etal-2018-mapping}. Each of these models depends on either human demonstrations \citep{misra-etal-2018-mapping, DBLP:journals/corr/abs-2107-01969} or predefined annotated tasks \cite{DBLP:journals/corr/abs-1912-01734} for replicability.

Moreover, different approaches have been adopted on a conglomerated scale to explore how agents collaborate in achieving predefined objectives. These have evolved into utilizing both the task description and the dialog interaction between the user and the agent as the language component. For example, in a Minecraft simulator, a collaborative building task involves an architect and a builder \cite{narayan-chen-etal-2019-collaborative}. Similarly, the IGLU (Improving Grounding and Language Understanding) task proposed by \citet{kiseleva2022interactive} models the behavior of both the architect and the builder separately. While the architect focuses on generating optimized and meaningful instructions, the builder aims to understand the instructions and operate within a simulator to complete the given task. Building upon the IGLU task, certain systems have been developed to establish collaborative two-way communication, enabling the builder to seek clarification from the architect \cite{mehta2023improving}.

\subsection{Multimodal transformers and Language Models}

Over the years, transformer-based language models have made significant strides in various tasks \cite{DBLP:journals/corr/VaswaniSPUJGKP17}. Multimodal transformer learning, focusing on integrating diverse modalities, has gained momentum over the years, allowing for an unprecedented synchronization range. Certain works use the transformer architecture to encode specific modalities \cite{DBLP:journals/corr/abs-2010-11929}. In the field of vision, numerous works have actively utilized language models and fusion techniques, closely linking them to contextually encoded pairs of multimodal data. For example, ViLBERT encodes visual and textual data, employing cross-attention at specific architecture layers \cite{DBLP:journals/corr/abs-1908-02265}. Similarly, LXMERT processes visual features and language data using cross-modal attention \cite{DBLP:journals/corr/abs-1908-07490}. In the context of embodied agents, a multi-transformer setting is used to sequentially process visual data guided by textual instructions \cite{DBLP:journals/corr/abs-2105-06453}.

\subsection{Language aspects of Embodied Agent Planning}

Improving the natural language communication abilities of embodied agents is crucial. Ideally, users should converse with the robot agent as they would with a person, with the robot efficiently extracting information and completing tasks. Additionally, integrating dialog acts that convey the communication intent of the conversation is beneficial \cite{gervits-etal-2021-agents}. This inclusion of dialog acts and their corresponding utterances aids in defining the overall plan required for specific task completion \cite{teachda}.

\subsection{LLMs as Embodied Agents Planner}

The current advancements in Large Language Models (LLMs) have led to various contributions in addressing embodied agents' challenges. These models leverage their extensive generalized knowledge to assist in this task. In their study, \citet{song2023llmplanner} explore the utilization of LLMs as planners for embodied agents operating in visually perceived environments. They harness the capabilities of large language models to perform few-shot planning for these agents. Similarly, in their work, \citet{zhang2023building} demonstrate the remarkable planning skills of LLMs in a single-agent embodied task. They also successfully illustrate how these acquired concepts can be applied in multi-agent scenarios where agents coordinate and communicate effectively in confined environments. Using LLMs can help expand the constraint environment, enabling the agent to become familiar with various scenarios \cite{wang2023voyager}.

\section{Dataset Description}

We utilize the \teach dataset \cite{padmakumar-etal-2022-exploring} in this task. The dataset is structured into various formulations, each aligning with a particular task. Essentially, it consists of dyadic conversations where human annotators assume the roles of users (\texttt{Commander}) and robots (\texttt{Follower}), working together on completing certain household tasks. In every dialog session, the \follower has a specific high-level task, such as \textsc{Make Plate of Toast} or \textsc{Prepare Sandwich}. The \commander possesses the details of this task, while the \follower does not. The task involves multiple sub-goals that the \follower must accomplish. To achieve this, the \commander conveys instructions to the \follower through dialogs. The \follower can backchannel or communicate with the \texttt{Commander}, seeking specific details and clarifications to enhance task efficiency. These clarifications often relate to inquiries regarding location, specific objects, or the steps to complete a task. The \follower now incorporates this collected data into a simulator to achieve its presumed goal. The data comprises around 3K recorded human-human gameplay sessions assimilated in an AI2-THOR virtual environment \cite{DBLP:journals/corr/abs-1712-05474}. 

In the \teach dataset, every session includes an initial state denoted as $S_i$ and a final state denoted as $S_f$. In addition, for a specific task, we have access to the aggregated subgoals, a sequence of actions performed by the \texttt{Follower}, represented as $A = (a_1, a_2, \ldots)$, and the dialogs between the \commander and \texttt{Follower}. Furthermore, the dataset offers synchronized visual frames extracted from the simulated environment in conjunction with the action sequence. For future reference, we denote the sequence of dialog actions as $A^{D}$ and all navigation and interaction actions as $A^{I}$.

\section{Task Formalisation}
\label{sec:task_formalisation}

\subsection{Execution from Dialog History (EDH)}
\label{subsec:edh}
In this task, we must automatically model how the \follower agent predicts actions to achieve a specific task or subgoal. 

An EDH instance is formed by taking a sub-sequence of a given gameplay session. We  define an EDH instance as $E_i = (S^E, A_H, A^I_R, I_H, F^E)$. Here,  $S^E$ is the initial environment state of the instance at time-step $t = 0$, and $F^E$ is the expected final environment state of the instance at $t = T_f$. The agent is given the action history $A_H$ along with \follower ego-centric image history $I_H$ till $t = T_i (0 \leq T_i \leq T_f)$. $E_i$ also comprises reference (ground-truth) interaction actions ($A^I_R$) for supervision. 

Let $\mathcal{A}$ be the set of all actions, and $\mathcal{I}$ be the set of \follower ego-centric images. The task here is to learn a function $f: \mathcal{A+I} \rightarrow \mathcal{A}$, that predicts the future sequence of actions $A_F$, which lead to the final environment state of the instance $F^E$. $A_F$ can be derived as:
\begin{center}
\begin{equation}
\label{eq:af}
    A_F = \{a_{T_i}, ..., a_{T_f}\}
\end{equation}
\begin{equation}
\label{eq:at}
    a_t = f(\{A_H+a_{t-1}\}, \{I_H+i_{t-1}\})
\end{equation}
\begin{equation}
\label{eq:it}
    i_t = simulator(a_t)
\end{equation}
\begin{displaymath}
    a_t \in \mathcal{A}, i_t \in \mathcal{I}, (T_i \leq t \leq T_f)
\end{displaymath}
\end{center}
Equation \ref{eq:it} represents the \follower ego-centric image in the simulator after processing action $a_t$. 

\subsection{EDH Plan Prediction}
\label{subsec:edh_plan_prediction}
\citet{teachda} defines task planning as generating a sequence of symbolic actions that guide a robot's high-level behavior to complete a task. A plan comprises a series of object manipulations that must be executed to achieve success. For instance, the robot may need to navigate to specific objects, and additional steps may be necessary depending on the environment's state. Despite this, generating a plan based solely on the dialog should be feasible. We extend the EDH task by predicting the future plans $P_F$ instead of future actions $A_F$.  

A reference plan $P_R$ is created by making $\{\text{action}, \text{object}\}$ pair from the \teach dataset \footnote{We keep the same splits as the original dataset} for every interaction action taken by the \follower and the object it interacts with, for a given EDH instance.  

EDH plan prediction task can be modeled as a sequence-to-sequence task, where given the dialog $A^D$ between the \follower and the \commander from an EDH instance as input, the agent is tasked to predict a plan $P_F$, which consists of all object interaction actions that happen during the EDH instance. The predicted plan $P_F$ consists of alternating object interaction actions $a_i$, $a_i \in A^O$ and object type $o_i$, $o_i \in O_R$ ($O_R$ is the set of all objects that the \follower can interact with in the simulator).

Let $\mathcal{P}$ be the set of all possible game plans. The task here is to fine-tune a model $g: \mathcal{A} \rightarrow \mathcal{P}$ that predicts a gameplan $P_F$ given the entire dialog $A^D$ for a given EDH instance.

\begin{equation}
    P_F = g(A^D)
\end{equation}

\section{Experiments}
\label{sec: experiments}
In this section, we present our methodological decisions to obtain conclusive results for the task described in Section \ref{sec:task_formalisation}. Since this task involves multiple modes, we evaluate the results and compare various training methods simultaneously to gain insights into the significance of each modality. Due to limited computational resources, we had to simplify our model configuration choices, which prevented us from experimenting with larger models and increased parameter sizes. The following sections elaborate on our adopted or tested approaches to better understand the selected data.

\subsection{Methodology}
This section describes the architecture used to achieve optimal configuration for multi-party embodied task completion. We have implemented a multi-transformer-based architecture similar to the one used by \citet{DBLP:journals/corr/abs-2110-00534} to address the issue of EDH. This allows us to track the path taken by the \follower using navigational and interactive actions. We investigate the generative strategy to predict game plans to validate the effectiveness of the language-only modality provided in the original dataset and demonstrate its versatility. This approach enables us to identify a combination of interactive actions and their corresponding object utilities that aim to achieve the goal of task completion.

\subsubsection{Episodic Transformer}
We utilize a multi-transformer architecture for the task of EDH (Execution form Dialog History), which involves predicting action sequence provided the dialog history ($A^{D}$) between the \commander and \texttt{Follower}, the visual camera observation taken from the AI2-THOR simulated household environment and the previous actions ($A^{I}_{H}$) taken by the \texttt{Follower} considering all the navigation and sub-task details. Our proposed architecture is primarily built upon the E.T. (Episodic Transformer) model \cite{DBLP:journals/corr/abs-2105-06453}.

Our architecture consists of four distinct modules: the BART AutoEncoder model \cite{DBLP:journals/corr/abs-1910-13461}, the Faster R-CNN \cite{DBLP:journals/corr/RenHG015} with a ResNet backbone \cite{DBLP:journals/corr/HeZRS15}, the Action Encoder, and a multimodal Cross-Attention module combined with a transformer Encoder block. These modules work together to produce likelihood scores for potential future action sequences. We opt for the BART language model to encode the dialog conversations. We choose this because BART excels in generating high-quality textual summaries and seamlessly integrates with dialog utterances. We encode the dialog history involving the \commander and the \follower and leverage the final hidden-state representation to represent the language modality. Next, we employ a ResNet-50 backbone to embed individual visual observations. Specifically, we utilize a Faster R-CNN trained ResNet-50 model for quicker computation of visual representations, primarily focusing on highlighting object regions within the images. Subsequently, we process these encoded visual representations through a Convolution block and apply linear transformations to ensure they are mapped to the same dimensional space as the language encoder. It should be noted that these representations are encoded independently of each other. We use the action encoder, a trainable embedding matrix, to map distinct action indices into a higher-dimensional space.

In contrast to the original E.T. architecture, we employ a cross-attention modeling technique that enhances hidden representations from various modalities by integrating contextual information from each other \cite{rajan2022crossattention}. To establish this cross-attention, we utilize the MultiHead Attention (MHA) module \cite{DBLP:journals/corr/VaswaniSPUJGKP17}. Unlike the conventional self-attention approach, which operates on similar representations for Query (\textit{Q}), Key (\textit{K}), and Value (\textit{V}), and calculates attention weights accordingly, in the cross-attention model, we assign a source modality to serve as both \textit{K} and \textit{V}, while a target modality functions as \textit{Q}. The underlying idea behind this approach is to explore cross-modal interactions by adapting the source modality to the target modality.
To ensure that this attention is not calculated over the future action indices during training, we construct an attention mask that enables the model to apply softmax over a specific indices position in the input.

Furthermore, it is important to note that we do not utilize the cross-modal MHA within a self-attentive framework. To accomplish this, we employ a multimodal encoder-block comprising multiple transformer-encoder layers \cite{DBLP:journals/corr/VaswaniSPUJGKP17}. After obtaining embeddings from modality-specific encoders and subsequent cross-modal attention ($h^{t}_{1:M}$, $h^{v}_{1:T}$, $h^{a}_{1:T}$), the multimodal encoder generates output embeddings ($z^{t}_{1:M}$, $z^{v}_{1:T}$, $z^{a}_{1:T}$). Causal attention is applied in the multimodal encoder to prevent visual and action embeddings from attending to future time steps.

We finally utilize the contextually aware hidden state output from the multimodal encoder, which reflects the visual observation indices. We then apply a linear transformation to independently classify actions and interactive object indices (See figure \ref{fig:et_plus_bart_architecture} for further reference).

\subsection{Additional Experiments}
\subsubsection{Dialog-Plan Grounding}
Additionally, we fine-tune models to predict future game plans within the EDH plan prediction task context. For this, we fine-tune LLaMA2 LLM \cite{DBLP:journals/corr/abs-2307-09288} and variants of BART on the processed version of the EDH dataset. Our selection of language model variants was driven by the need to explore a range of models with varied sizes and capabilities.
    
We work on refining multiple BART LM versions, including the base and large variants. Our primary focus revolves around understanding the performance variations among the models based on parameter size for the dialog-to-plan generation task. Furthermore, we train the BART-large model using synthetic language data \cite{DBLP:journals/corr/abs-1912-01734} accessible along with the \teach benchmark dataset.  To assess its adaptability and the potential benefits of synthetic data in improving results, we refine it again on the current dataset.
 
Finally, we conduct experiments using LLaMA2 LLM because of its specialization in dialog understanding and generation. Moreover, we aim to emphasize the performance superiority inducted by utilizing LLM and its overall efficiency in leveraging the given task.

\subsubsection{MAF-BART}
We conduct experiments to predict future action sequences in the EDH (Execution from Dialog History) task. The prediction relies on input data, including dialog history ($A^{D}$), prior actions ($A^{I}_{H}$), and synchronized visual camera observations. Overall, the technique under consideration resembles language generation, where the output at each time step ($t$) depends on the encoded input information and the previously generated output. To achieve this, we develop a modeling architecture based on a BART LM and integrate the Multimodal Aware Fusion (MAF) \citet{kumar2022did} module to seamlessly incorporate contextual knowledge from external modalities. MAF employs a Multimodal Context-Aware (MCA2) attention \cite{DBLP:journals/corr/abs-1902-05766}, to efficiently combine textual and multimodal information, resulting in enhanced predictive capabilities for future actions. Finally, we utilize a decoder block constituting transformer-decoder layers \cite{DBLP:journals/corr/VaswaniSPUJGKP17} to predict future actions accordingly. For further detail relative to the architecture, refer to section \ref{appendix:maf}.

\subsubsection{Training \& Inference}
We conduct training for the tasks and configuration mentioned above using a system equipped with Nvidia A100 GPU 80GB. However, we encountered challenges related to certain driver issues and the AI2-THOR simulator's incompatibility with the remote cluster hosting the GPU. Consequently, we were unable to carry out inference. As a result, we had to limit access to most of our model configurations, making them compatible with a smaller machine equipped with an Nvidia 1050Ti GPU 4GB. This led to the removal of several high-end model configurations, as they were hindering the compliance of the vision modality to the local machine.

In this section, we will outline the various hyperparameter settings that we experimented with and provide supplementary meta-data information regarding the proposed modeling techniques and attaining improved results for the selected \teach dataset. 

In the context of Execution from Dialog History (EDH), we explore various modeling approaches within the E.T. model architecture and analyze their specific purposes through ablation experiments. However, on a broader scale, we opt for a Faster-R-CNN pretrained ResNet-50 model to represent visual observations, with each frame $\text{f} \in R^{(512 \times 7 \times 7)}$. We utilize a GPLM, namely the BART-base\footnote{\url{https://huggingface.co/facebook/bart-base}} model to encode language instruction ($A^{D}$). BART contextually encodes each token in the language input to $R^{768}$ dimensional vector space. Furthermore, we conduct experiments using a fine-tuned version of the BART model on the Synthetic dialog-action dataset provided along with the original \teach dataset. To conglomerate multiple modalities, we transform them into a common subspace of $R^{768}$. Next, to facilitate the contextual self-attentive fusion of multimodal features from each modality, we select an encoder block with 2 transformer-encoder layers\footnote{\url{https://pytorch.org/docs/stable/generated/torch.nn.TransformerEncoder.html}}. In the context of the EDH task, we receive a clear separation between history (previously executed) and future actions. However, during training, we combine them into sequences of visual observations and actions, with the option to calculate loss that includes history time steps. 

We finally execute dual linear classification on every visual index representation obtained from the multimodal encoder to categorize actions and their subsequent objects. We apply CrossEntropy loss to each classification layer and then aggregate and backpropagate them as a sum. Table \ref{tab:hyperparam_et} presents the selected hyperparameter attributes for this task.

Furthermore, to incorporate the impact of synthetic language simplification, we refine the BART-base model using synthetic language data\footnote{\url{https://huggingface.co/Koshti10/BART-base-ET-synthetic}}. This data aligns task descriptions within the ALFRED dataset \cite{DBLP:journals/corr/abs-1912-01734} and associates them with action-oriented simplification similar to machine-generated high-level code. We conduct the model fine-tuning in a conditional generation setup over 20 epochs. Overall, approximately it takes around 20 hrs to extract features, train and infer a model variant.

Within the context of the EDH plan prediction task, we used Colab \footnote{\url{https://colab.research.google.com/}} with Nvidia A100 GPU (80GB) for fine-tuning each of the 4 modeling strategies. Key hyperparameters for BART and LLaMA2 LLM are provided in table \ref{tab:finetuning-hyperparameters}. To fine-tune a very large model like LLaMA2 LLM within the given runtime constraints, we used QLoRA \citet{dettmers2023qlora}, which backpropagate gradients through a frozen, 4-bit quantized pre-trained language model into Low-Rank Adapters.

\section{Result \& Analysis}

\begin{table}[!htb]
\resizebox{\columnwidth}{!}{%
\begin{tabular}{p{2.5cm}p{2cm}p{2cm}}
\hline
\multirow{2}{*}{Model} & \multicolumn{2}{c}{\textbf{Valid Seen}} \\ \cline{2-3} 
                                & SR [TLW]  & GC [TLW]  \\ \hline
\multicolumn{3}{c}{\teach Baseline}                              \\ \hline
Lang                            & \hspace{0.5mm}0.72 [0.11]        & \hspace{2.5mm}0.89 [0.15]       \\
ET                              & \hspace{0.5mm}3.66 [0.72]        & \hspace{2.5mm}6.82 [1.47]       \\
+ H                             & \hspace{0.5mm}5.78 [0.97]        & \hspace{2.5mm}8.60 [1.99]       \\
+ S                             & \hspace{0.5mm}7.22 [1.78]        & \hspace{0.5mm}10.93 [2.96]      \\
+ H + S                         & \hspace{0.5mm}7.45 [1.34]        & \hspace{0.5mm}12.59 [3.80]      \\ \hline
\multicolumn{3}{c}{\textbf{Ours}}                                         \\ \hline
Lang                            & \hspace{0.5mm}1.99 [0.45]        & \hspace{2.5mm}3.14 [1.12]       \\
ET + BART                       & \hspace{0.5mm}5.08 [0.63]        & \hspace{2.5mm}7.34 [1.28]       \\
+ H                             & \hspace{0.5mm}6.85 [1.21]        & \hspace{2.5mm}9.67 [2.43]       \\
+ S                             & \hspace{0.5mm}7.12 [1.46]        & \hspace{0.5mm}10.33 [3.39]      \\
+ H + S + CA                    & \hspace{0.5mm}8.85 [1.09]        & \hspace{0.5mm}14.02 [3.72]      \\ \hline
\end{tabular}%
}
\caption{The table presents results for different modeling strategies on evaluation metrics, including success rate (SR), goal-conditioned success rate (GC), and trajectory length weighted metrics (in brackets). We perform ablations on history loss (H), language encoder initialization with ALFRED synthetic language (S), and the use of cross-modal attention before fusion (C). }
\label{tab:edh_results}
\end{table}

\begin{table}[!htb]
\centering
\resizebox{\columnwidth}{!}{%
\begin{tabular}{p{2.5cm}p{1.15cm}p{1.15cm}p{1.15cm}}
\hline
\multirow{2}{*}{Model} & \multicolumn{3}{c}{\textbf{Valid-Seen}}           \\ \cline{2-4} 
                       & R-1 & R-2 & R-L  \\ \hline
BART                   & \hspace{1.5mm}0.59  & \hspace{1.5mm}0.00     & \hspace{1.5mm}0.58      \\ 
BART-base              & 38.93 & 28.83 & 35.73    \\ 
BART-large             & 37.99 & 27.64 & 34.81      \\ 
BART-synthetic         & 38.22 & 27.74 & 35.43     \\ 
LLaMA2 LLM                 & 54.14 & 41.22 & 46.77    \\ \hline
\end{tabular}%
}
\caption{The table compares different fine-tuned models on ROGUE-1 (R-1), ROGUE-2 (R-2) and ROGUE-L (R-L) metrics. The Vanilla BART-base model (without any fine-tuning) is selected as the baseline}
\label{tab:finetune}
\end{table}

\subsection{Evaluation Metrics}
\label{subsec:eval}
This section lists various evaluation metrics that assess the modeling strategies discussed in Section \ref{sec: experiments}, matching different task definitions and functionalities. 

EDH task is evaluated on 2 metrics for each  instance, along with a trajectory length weighted score for each metric:
\begin{itemize}
    \item Success: Success is 1 if \emph{all} expected actions $A^I_R$ are present in $A_F$, else 0. Success Rate is calculated by averaging over all instances\footnote{Authors use macro-average instead of micro-average}. 
    \item Goal-Condition Success: Fraction of expected actions $A^I_R$ present in $A_F$. Goal-Condition Success Rate is calculated by averaging over all instances.
    \item Trajectory Weighted Metrics:  This is calculated for each of the above two metrics for a metric value $m$ as follows :
    \begin{displaymath}
        \mathrm{TLW}_{m} = \frac{m \times |A^I_R|}{\mathrm{max}(|A^I_R|, |A_F|)}
    \end{displaymath}
\end{itemize}

To assess the effectiveness of the predicted game plan, we employ ROUGE scores in a sequence-to-sequence task. ROUGE evaluates the quality of the generated text by analyzing word and phrase overlaps with reference dialog conversations. This approach ensures that the output is both linguistically accurate and contextually appropriate. We utilize the ROUGE-1, ROUGE-2, and ROUGE-L to evaluate our models, thus considering the unigram, bigram, and longest-subsequence performance into account.

\subsection{Quantitative Analysis}
\subsubsection{EDH (Execution from Dialog History)}
Through table \ref{tab:edh_results}, we present a comprehensive evaluation of our proposed approach using three distinct metrics: success rate (SR), goal-conditioned success rate (GC), and trajectory length-weighted metrics (TLW). Our modeling strategies consistently outperform the baseline models. Our approach leverages a BART language model to encode dialog history ($A^{D}$), setting it apart from the transformer-based encoder block used in the baseline models. Additionally, it is evident from the table that the performance of the employed architecture is subpar and that the model cannot be relied upon for real-life applications as an embodied agent system. However, the implemented changes still yield reasonable results for certain tasks, demonstrating their relative reliability in specific subsamples of \teach dataset.

The performance of the ablated baseline E.T. model in the original study by \citet{DBLP:journals/corr/abs-2110-00534} may differ from the ones mentioned in the table. This is because the original study utilized a bigger variant of the training dataset, which has now been reduced to only 5K instances. As a result, the number of task-specific instances has also decreased. To compare our proposed approach, we recreated the baseline models with a similar setup described in the original study.

Firstly, the table clearly shows that the model's performance is below average when it receives only language-based information, i.e., the \commander and \follower dialog interactions. This is because the model faces difficulty in dealing with lengthy sequences of navigational actions and infrequent interactive behaviors, as these largely depend on the camera observations from preceding actions. Consequently, the model that relies solely on language cannot provide enough contextual information to influence its predictions in the long run. Moreover, since we use the BART language model, which excels at summarizing dialog conversations, we achieve better-aligned performance compared to a regular transformer-based language encoder with a vocabulary solely focused on dataset-oriented words/tokens. 

Incorporating visual observations into the data modeling is crucial, as indicated in the table where there is a significant improvement in the outcomes.  Specifically, when evaluating the success rate based on goal conditions, which assesses the fractional alignment of predicted output actions with the ground-truth samples, we observe a substantial increase in GC values for both the baseline E.T. model and our model. Nevertheless, owing to the significance introduced by the BART encodings, we achieve a marginally increased score on the validation examples for this metric. Furthermore, considering the trajectory-length weighted values for both metrics in the present scenario, we can see that our model prioritizes predicting shorter trajectory lengths compared to the expected path. Thus indicating a higher classification precision per instance across the validation split of \teach dataset.

Our analysis of model architectures that consider the complete history of action subsequences for calculating cross-entropy loss during training yielded noteworthy improvements. Considering the definitive classification of whole subsequences of actions, we see a substantial 2\% increase in the success rate (SR) and improved trajectory length weighted (TLW) values. Moreover, this trend also applies to the goal-conditions success rate (GC), where we could classify almost one-tenth of the goal conditions for each instance in the validation set.

We observe a slight decline in the performance of our E.T. model when compared to the baseline. Our E.T. model uses the fine-tuned BART model on a synthetic language dataset, achieving a predicted ROUGE-LSum score of 75\%. Consequently, the reason for this dip in performance can be attributed to the usage of synthetic language encoder weights from the original work, as instructed by \citet{DBLP:journals/corr/abs-2105-06453} in training the baseline E.T. model (replication). It should be noted that the original work train their model on both the ALFRED dataset and the synthetic dataset simultaneously, owing to a one-to-one mapping of instances. However, we only train our model on the \teach dataset, without external influence from ALFRED data, thus showing a correlation with the depletion of scores.

Finally, to demonstrate the efficacy of our approach, we take into account all modeling variations, including the implementation of synthetic training to standardize task descriptions, the utilization of interactions between the three modalities ${\text{language}, \text{frames}, \text{actions}}$ via cross-modal attention, and the consideration of history loss. Under this setting, we attain a higher success rate (SR) and goal-conditioned success rate (GC). To elaborate on the GC, we observe a notable 4\% improvement in effectively classifying action sequences in discrete goal-condition cases. Nevertheless, it is essential to highlight that the TLW, specifically concerning SR, has significantly diminished. This implies that accurately forecasting positive SR invariably involves an extended trajectory length in action sequences.

\subsubsection{EDH Plan Prediction}
In table \ref{tab:finetune}, we assess the performance of the fine-tuned model configurations on the validation set prepared for the task. In this context, we utilize ROGUE scores, which are discussed in subsection \ref{subsec:eval}. Our baseline choice involves opting for a zero-shot BART-based model without external training. This approach enables us to assess how fine-tuning and model improvements affect performance.

The vanilla BART model lacks prior knowledge about the game plans, resulting in a generative response that does not include general information about any game plans or actions. Nevertheless, when we fine-tune the BART-base model, we observe performance improvement, as indicated in the table. Moreover, The model learns and recognizes task-related cues and context within dialogs, allowing it to generate plans closely aligned with the intended tasks. Furthermore, it consistently maps similar dialogs to the same or similar plans, improving task consistency and resulting in more reliable task predictions.

A  similar behavior is observed when considering the fine-tuned BART-large model on the game-plan prediction dataset. However, this occurs because the original EDH dataset contains many recurring instances of similar gameplay across multiple instances. Consequently, the resulting dataset lacks variety in distinct plan dialogs. The reason for this and certain model initialization factors is that we observe a marginal degradation in performance.

Integrating the LLaMA2 LLM model into the task, as formalized previously, presents an intriguing prospect for improving the performance and addressing the challenges associated with this task compared to using BART models. LLaMA models are designed to understand and generate human language with fluency and context awareness. This enhanced language understanding capability is particularly valuable for comprehending nuanced and context-rich dialogs in the task. The longer context length of LLaMA allows it to capture more extensive conversational history, thus better capturing relevant information.

\subsection{EDH with Generative Action Decoding}
Our attempts to transform the EDH (Execution from Dialog History) task into a generative problem, wherein the actions for each subsequent time step are decoded using the presented multimodal input data, yielded inconclusive results. Due to the computational constraint at inference, we restrict our model from camera observation aligning with future actions. Moreover, we completely rely on the forecasting ability of the model to generate a response.

We assess the proposed strategy using the F1 metric, which necessitates predicting the output sequence alongside the ground-truth values. Our F1-score is relatively low at $0.3562$, primarily due to the significant overrepresentation of navigational action indices in the entire EDH dataset. The substandard performance can be attributed mainly to the utilized framework. Despite the potential benefits of the MAF strategy in contextualizing modalities, the visual data fails to synchronize with the discourse, contrary to SED dataset performance from the original work by \citet{kumar2022did}. As a result, even with recurrent training, the model can not extract meaningful insights from the processed data.

\subsection{Qualitative Analysis}

Table \ref{tab:train_meta_data_teach} offers a general statistical summary of the task-specific information in the \teach dataset. In general, the dataset presents gameplay sessions with varying task complexities, each demanding different cognitive skills for execution by the follower. Certain tasks require the \commander and \follower to engage in fewer dialog interactions and involve fewer variable parameters in the form of objects for completion. Nonetheless, we can also affirm that the provided dataset exhibits a greater prevalence of navigation action sequences. 

Regarding the EDH dataset, we can demonstrate an excessive recurrence of navigational actions within the ${A_H, A^{I}_{R}}$ subsequences. To be precise, the navigational actions, \texttt{Forward}, \texttt{Pan Right}, \texttt{Pan Left}, \texttt{Turn Left}, \texttt{Turn Right} make up approximately 80\% of the total actions executed by the \follower in the task dataset. The table clearly illustrates that, for certain specific tasks, object-specific interactive actions make up only one-tenth of the total actions needed to complete the task. This is illustrated through figure \ref{fig:ex_1_edh} sampled from the train-split of the EDH dataset. Here, the task is to execute the following: i) \textsc{N Slices of X in Y}, and ii) \textsc{Put All X in One Y}. However, this requires 138 actions to be executed, most of them being consecutive navigational actions.

Generally, the model predicts and executes particular actions in real time. We supply it with language instructions, action history, and visual observations, enabling the model to establish a connection between the given instructions and the subtasks necessary for task completion. Nevertheless, this acquired characteristic of the model obstructs the overall task execution. Additionally, the model tries to quickly identify the crucial step needed to advance a specific subtask but, in doing so, neglects the intricate smaller steps required for successful completion, which accumulates over time and impacts the later stages. The problem is illustrated through figure \ref{fig:ex_2_edh}. In this, the task is to \textsc{Make Sandwich}. It constitutes of the following subtasks, i.e., \{{\textsc{Pickup Knife}, \textsc{Slice Bread}, \textsc{N Slice of Bread in Plate}, \textsc{Slice Tomato}, \textsc{N Slice of Tomato in Plate}\}, however, the model fails to pick up either item and place them into to the plate. Consequently, when the final state of the simulator is observed, it could not identify sliced bread or sliced tomato, thus resulting in task failure, with 2 GC satisfied.

According to our analysis, we realize that the authors of the original \teach benchmark dataset aim to replicate human behavior in AI agents. To achieve this, they use human-human game roleplay recordings where participants can perform tasks in distinctive ways, explicitly following specific predefined instructions. The fundamental requirement for task completion involves measuring the initial state $S_{I}$ and final state $S_{F}$ of the simulated environment using metrics like SR and GC. However, this substantially impacts the overall performance of the acquired embodied agent. The differences in execution strategies employed during model learning for each instance become apparent, highlighting variations in individuals' approaches toward tasks. Furthermore, the limited number of unique instances per task poses a challenge for effective generalization, resulting in the model making errors in its decisions.

The model excels in predicting the short sequence of actions for a specific task and is invariant with additional navigational details. Particularly when the reference frame remains nearly consistent and the model needs to carry on with the task. Next, we demonstrate the inefficiency exhibited by the \teach benchmark dataset and outline flaws in the current evaluation system. In the context of the AI2-THOR simulated environment, glitches have been observed to occur in various scenarios in a stochastic manner. For example, in Figure \ref{fig:ex_4_edh}, we see the task of \textsc{Clean Coffee Mug}. The model accurately forecasts the subsequent actions of \textsc{Pour in Sink}. Nevertheless, because of specific internal technical problems, it cannot complete the task, resulting in incorrect outcomes later. 

In Figure \ref{fig:ex_3_edh}, we observe an EDH instance that illustrates a 100\% success rate (SR) and a goal-conditioned success rate (GC). However, upon closer examination, we discern that the modeling agent skips the step of \textsc{Clean Pot} despite being assigned the task of \textsc{Boil Potato}. Furthermore, the evaluation process relies heavily on locational information and attributes objects based on their final state in the environment. Through their work, the authors failed to include the essential practice of assessing the state of objects under task consideration during the evaluation.

\section{Conclusion \& Future Work}

In this work, we present a multi-transformer-based system that leverages a BART GPLM and employs a cross-modal attention strategy to learn the task-solving capabilities of an embodied agent. Despite the apparent inadequacy of the task performance and its unsuitability for practical application, we propose a deduction to enhance the data quality to achieve greater generalizability. We discuss the the importance of the language modality and the aspects of utilizing it to model a game plan, which serves as the high-level perspective of a task. We demonstrate the ability of LLMs as planners to generalize and induce human-oriented knowledge to execute tasks based on dialog interactions. However, because of specific computational limitations, we cannot achieve the optimal modeling configurations that are feasible with the dataset under consideration. 

We propose improvements in modeling strategies and data formulation to enhance current performance. First, we recommend experimenting with a model like the Vision Transformer to effectively represent camera observations because the current model sometimes struggles with the abstract quality of the image. Furthermore, we propose aligning image representation with dialog interactions based on the insights learned from the unsuccessful experiments with the MAF-BART model. Posing the EDH task as a topological classification strategy could also negate the influence of navigational movements (imbalance). The use of multimodal LLMs could be exploited to tackle the issue of embodied agents under a few-shot setting. To enhance the performance of the mentioned models, we can implement soft rules to limit the recording of multi-party gameplay and carefully curate a larger dataset.

\bibliography{anthology,custom}

\appendix

\section{Appendix}
\label{sec:appendix}

\begin{table}[!htbp]
\resizebox{\columnwidth}{!}{%
\begin{tabular}{p{3.65cm}p{3.4cm}}
\hline
Hyperparameter      & \hspace{9mm}Value         \\ \hline
epochs              & \hspace{9mm}20            \\
batch size          & \hspace{9mm}2             \\
loss                & \hspace{9mm}Cross-Entropy \\
optimizer           & \hspace{9mm}AdamW         \\
learning rate (lr)  & \hspace{9mm}1e-5          \\
lr scheduling type  & \hspace{9mm}Linear        \\
lr decay rate       & \hspace{9mm}0.1           \\
weight decay        & \hspace{9mm}0.33          \\
$\mathrm{action\_loss}$ weight & \hspace{9mm}1             \\
$\mathrm{object\_loss}$ weight & \hspace{9mm}1             \\
dropout rate        & \hspace{9mm}0.2           \\ \hline
\end{tabular}%
}
\caption{Displays the selected hyperparameter settings for training the E.T. model in the context of the EDH task. These attributes remain fixed for all modeling configurations related to this task, ensuring that the results obtained are invariant with these values.}
\label{tab:hyperparam_et}
\end{table}

\begin{table}[!htb]
\resizebox{\columnwidth}{!}{%
\begin{tabular}{p{3.45cm}p{3.25cm}}
\hline
\multicolumn{2}{c}{BART} \\ \hline
Hyperparameter & Value \\ \hline
epochs & 20 \\ 
batch size & 8 \\ 
learning rate & 5e-5 \\ 
warmup steps & 500 \\ 
weight decay & 0.01 \\ \hline
\multicolumn{2}{c}{LLaMA2 LLM} \\ \hline
Hyperparameter & Value \\ \hline
epochs & 1 \\
batch size & 4 \\
learning rate & 2e-4 \\
weight decay & 0.001 \\
LORA alpha & 16 \\
LORA dropout & 0.1 \\
LORA r & 64 \\ \hline
\end{tabular}%
}
\caption{Displays the hyperparameter settings for finetuning BART models and LLaMA2 LLM model}
\label{tab:finetuning-hyperparameters}
\end{table}

\begin{figure*}[!htb]
    \centering
    \includegraphics[width=\textwidth]{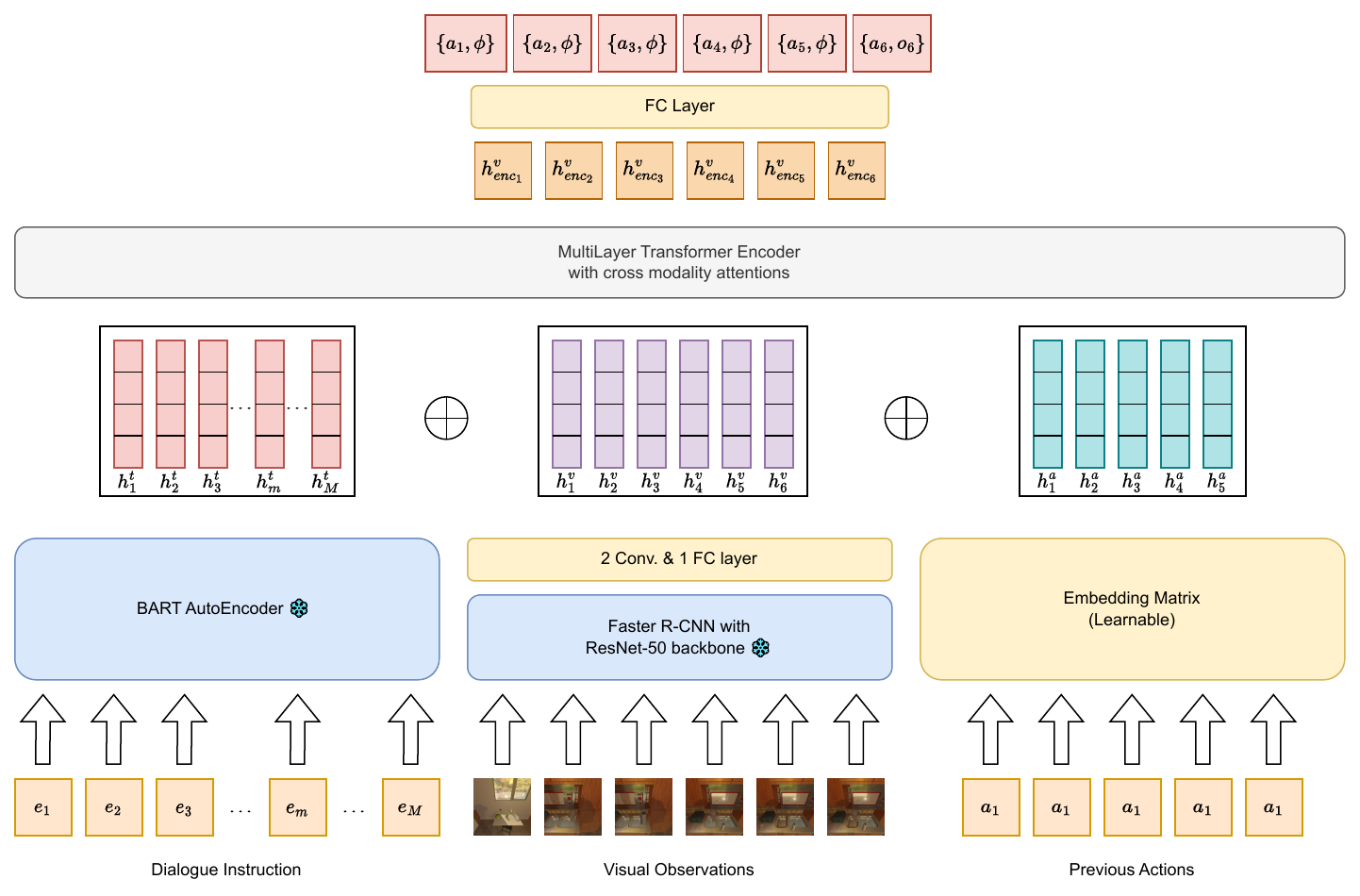}
    \caption{The figure depicts the Episodic-Transformer architecture, which incorporates the BART language model and a multimodal cross-attention module. This module enables the contextual transfer of information across different modalities. In the illustration, we observe an inference scenario involving the dialog history $e_{1:M}$ (\textsc{Lang}), visual observations from the start of the episode, denoted as $v_{1:t}$ (\textsc{Frames}), and previously executed actions $a_{1: t-1}$ (\textsc{Actions}). The model operates by processing $e_{1:M}$ using a BART AutoEncoder, transforming $v_{1:t}$ using a ResNet backbone, and embedding $a_{1:t}$ via an embedding matrix. It then generates actions and, if applicable, interactive objects at the current time step, $t$.}
    \label{fig:et_plus_bart_architecture}
\end{figure*}

\begin{table*}[!htb]
\resizebox{\textwidth}{!}{%
\begin{tabular}{p{6.15cm}p{2.15cm}p{3.25cm}p{3.25cm}}
\hline
                                   & \begin{tabular}[c]{@{}l@{}}Total\\ Sessions\end{tabular} & \begin{tabular}[c]{@{}l@{}}Average Actions\\ per Session\end{tabular} & \begin{tabular}[c]{@{}l@{}}Average Interactive\\ Actions per Session\end{tabular} \\ \hline
\textsc{Water Plant}               & 91                                                       & $68.27 \pm 44.29$                                                     & $7.05 \pm 6.69$                                                                   \\
\textsc{Plate Of Toast}            & 80                                                       & $139.80 \pm 81.81$                                                    & $19.80 \pm 10.46$                                                                 \\
\textsc{Clean All X}               & 169                                                      & $98.30 \pm 82.14$                                                     & $17.27 \pm 14.82$                                                                 \\
\textsc{Sandwich}                  & 89                                                       & $255.19 \pm 116.28$                                                   & $41.76 \pm 13.06$                                                                 \\
\textsc{N Slices Of X In Y}        & 144                                                      & $145.64 \pm 121.01$                                                   & $19.44 \pm 14.66$                                                                 \\
\textsc{Coffee}                    & 145                                                      & $69.94 \pm 46.44$                                                     & $10.55 \pm 6.71$                                                                  \\
\textsc{Put All X In One Y}        & 163                                                      & $158.19 \pm 115.81$                                                   & $9.57 \pm 7.86$                                                                   \\
\textsc{Boil X}                    & 81                                                       & $126.74 \pm 96.09$                                                    & $18.70 \pm 16.12$                                                                 \\
\textsc{Salad}                     & 139                                                      & $258.13 \pm 107.24$                                                   & $40.37 \pm 14.14$                                                                 \\
\textsc{Breakfast}                 & 113                                                      & $340.59 \pm 151.04$                                                   & $56.30 \pm 18.81$                                                                 \\
\textsc{Put All X On Y}            & 182                                                      & $101.73 \pm 83.48$                                                    & $8.25 \pm 6.40$                                                                   \\
\textsc{N Cooked Slices Of X In Y} & 89                                                       & $185.70 \pm 96.56$                                                    & $31.69 \pm 13.52$                                                                 \\ \hline
\end{tabular}%
}
\caption{showcases the meta-data information for the total gameplay session available across the training split of the \teach dataset. (Left) Displays all the unique tasks found in the dataset. For each task, we show the average total actions per session and the average total interactive actions per session, along with their respective deviations in values.}
\label{tab:train_meta_data_teach}
\end{table*}

\begin{table*}[!htb]
\resizebox{\textwidth}{!}{%
\begin{tabular}{p{12cm}p{5cm}}
\hline
Task Description                                                                                                                                                                                                                                                                                                                                                                                                    & \begin{tabular}[c]{@{}l@{}}Synthetic Language \\ Simplification\end{tabular} \\ \hline
Turn right toward the towel ring on the wall. Pick up the towel from the towel ring. Turn around and head to the toilet. \textcolor{red}{Put the towel on the back of the toilet}. Turn to the right, go toward the tub, and turn right. Pick up the towel from the ring on the wall. Turn around and go back to the toilet. \textcolor{red}{Put the towel on the back of the toilet} to the left of the bottle.                                      & Put two towels on the back of the toilet                                     \\ \hline
Turn around and \textcolor{red}{go to the sink}. Grab the \textcolor{red}{white lotion bottle with a brown pump}. Turn around and back up towards the bathtub, then turn around to the sink again. Open the right \textcolor{red}{cabinet door under the sink} and place the lotion inside.                                                                                                                                                                                                     & Place the lotion on the sink underneath the cabinet.                        \\\hline
Turn and go to the counter by the sink. Pick up the knife on the counter. Cut the tomato on the counter.  Turn and go to the microwave. Put the knife in the microwave. Turn and go to the counter by the sink. Pick up a \textcolor{red}{slice of tomato} on the counter. Turn and go to the refrigerator.  \textcolor{red}{Chill the slice of tomato in the refrigerator}. Turn and go to the microwave.  \textcolor{red}{Put the slice of tomato in the microwave.} & Put a chilled slice of tomato in the microwave.                              \\ \hline
\end{tabular}%
}
\caption{Showcase, instance sampled from the synthetic dataset provided through the ALFRED dataset \cite{DBLP:journals/corr/abs-1912-01734}.}
\label{tab:synthetic_data_ex}
\end{table*}

\begin{figure*}[!htb]
    \centering
    \includegraphics[width=\textwidth]{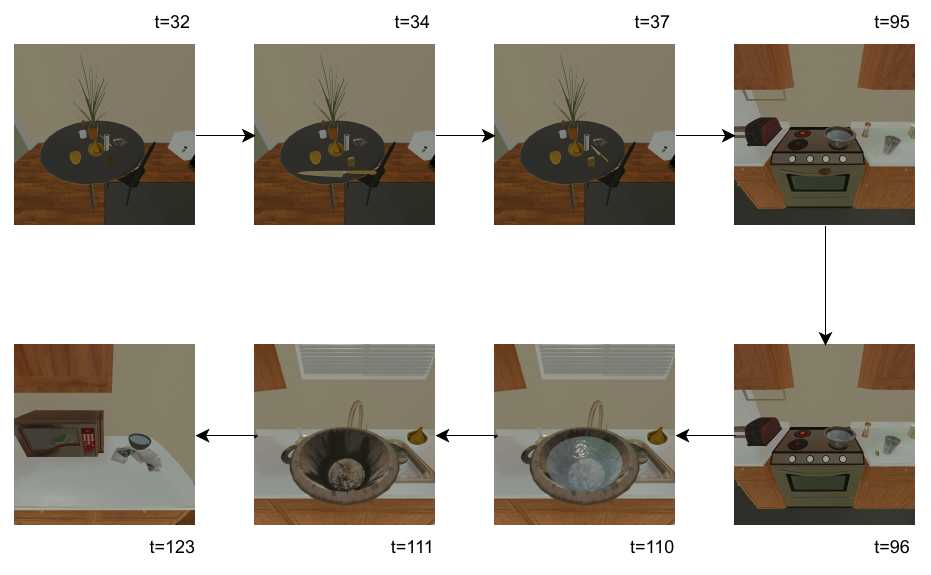}
    \caption{Shows the \follower ego-centric camera observation from all of the available interactive action timesteps of an EDH task instance. The task involves slicing a potato and placing the slices into an empty pot. The suggested total number of executable actions in the subsample is 138.}
    \label{fig:ex_1_edh}
\end{figure*}

\begin{figure*}[!htb]
    \centering
    \includegraphics[width=\textwidth]{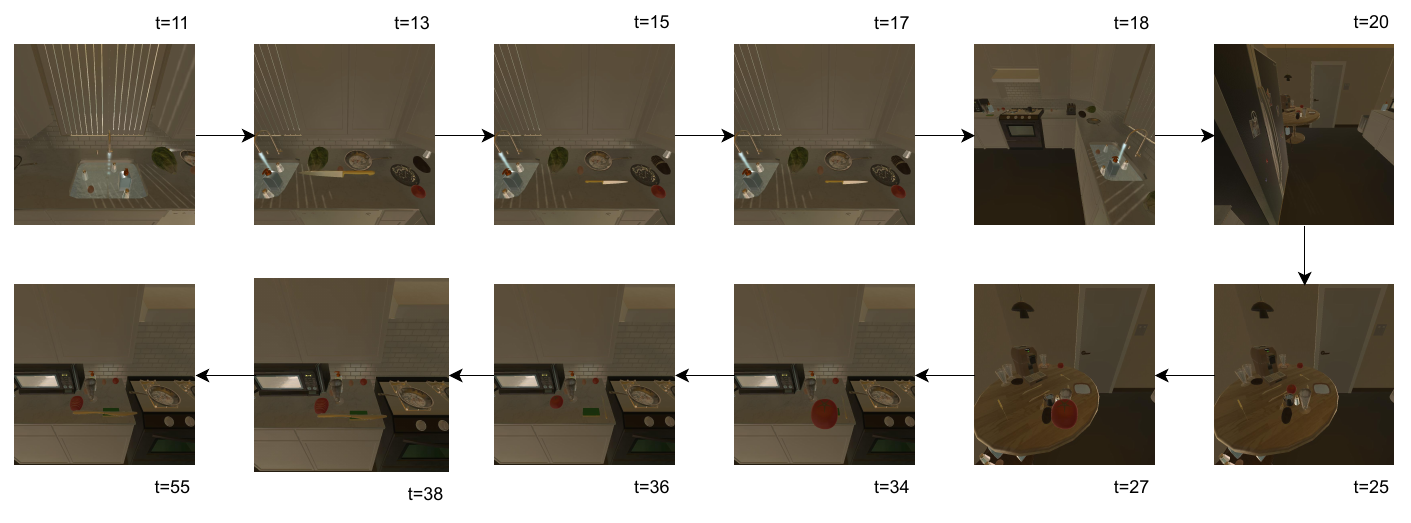}
    \caption{Illustrates the ego-centric camera observations of the \textsc{Make Sandwich} task, highlighting interpretative actions carried out at each time step that are essential for completion. Nevertheless, starting from t=38, the model struggles to correlate visual perception with the task instructions, resulting in a continuous loop.}
    \label{fig:ex_2_edh}
\end{figure*}

\begin{figure*}[!htb]
    \centering
    \includegraphics[width=\textwidth]{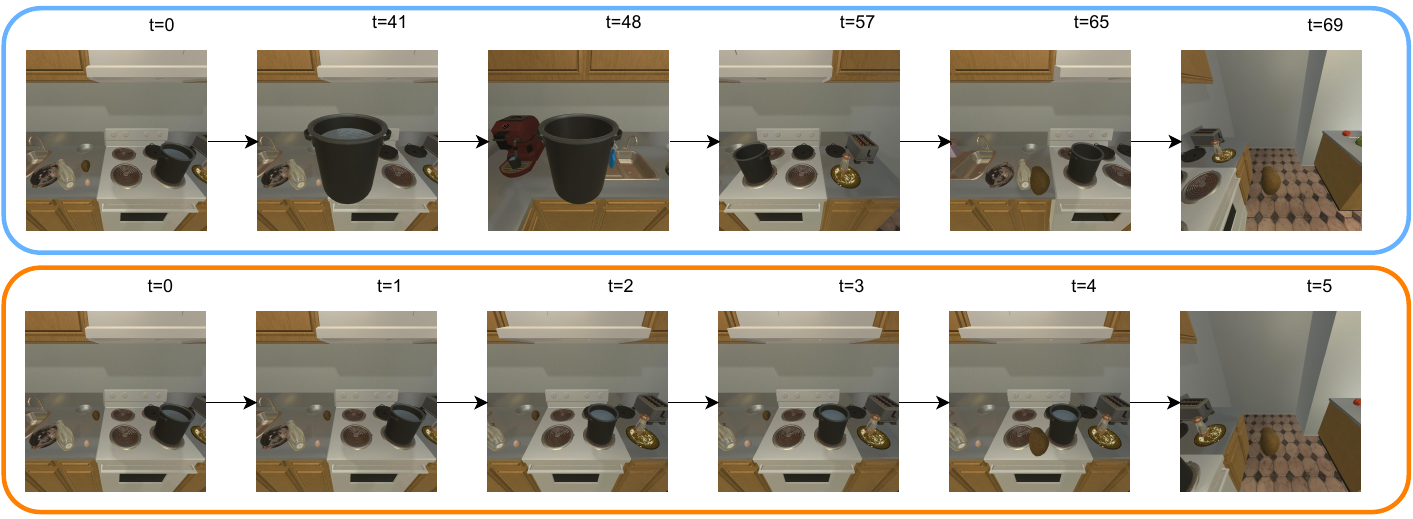}
    \caption{Showcases an EDH instance and compares the generated response against the ground-truth action sequences. It highlights the shortcomings of the employed evaluation strategy.}
    \label{fig:ex_3_edh}
\end{figure*}

\begin{figure*}[!htb]
    \centering
    \includegraphics[width=\textwidth]{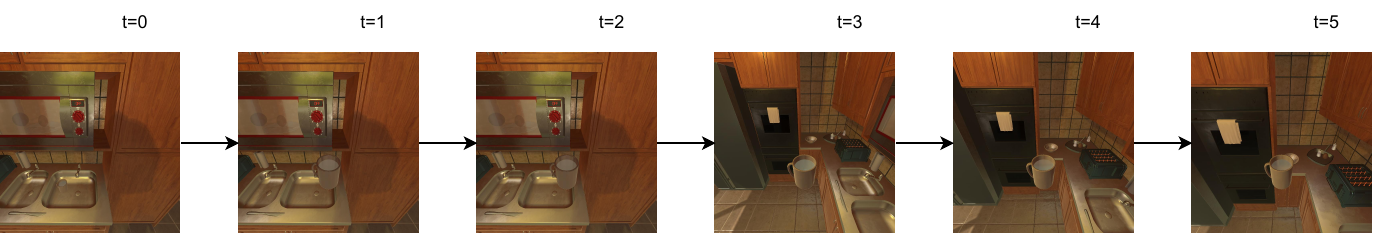}
    \caption{Shocases a positive EDH instance of executing the task of \textsc{Clean Coffee Mug}.}
    \label{fig:ex_4_edh}
\end{figure*}

\subsection{MAF-BART (extended)}
\label{appendix:maf}

We conduct experiments involving the prediction of future action sequences within the context of the EDH (Execution from Dialog History) task. This prediction relies on input data consisting of dialog history ($A^{D}$), prior actions ($A^{I}_{H}$), and synchronized visual camera observations, all of which influence the outcome. This task is similar to language generation, where the output at time-step $t$ depends on both the encoded input information and the previously generated output in the sequence (See equation \ref{eq: cond_gen} for further reference).

\begin{equation}
P(X)=\prod_{t=1}^M P\left(a^{f}_t \mid a^{f}_{1:t-1}, X_{in}\right)
\label{eq: cond_gen}
\end{equation}

\begin{equation}
    X_{in} = \mathcal{F}\left(A^{D}, A^{I}_{H}, V_{H}\right)
\end{equation}

Here, $X_{in}$ represents the encoded information obtained from modeling the input data. This information is subsequently used to condition future action predictions.

We construct our model architecture based on a BART language model, which incorporates encoder-decoder models to enable conditional generation adaptability. To seamlessly integrate multimodal knowledge into the BART architecture, we employ a Multimodal Aware Fusion (MAF) module, similar to the one described by \citet{kumar2022did}. The MAF module is adapter-based and consists of Multimodal Context-Aware (MCA2) \cite{DBLP:journals/corr/abs-1902-05766} Attention and Global Information Fusion (GIF) mechanisms. We use this module to incorporate contextual knowledge from external modalities into the source-text representations when given textual input in the form of pre-tokenized dialog history $A^{D}$ and a sequence of action-frame cues.

\textbf{Multimodal Aware Fusion}, the standard cross-modal attention in the form of dot-product relies on the direct interaction between textual representations and other modalities. However, because modality representations may not have the same subspace dimensionality as textual representations, this can result in information loss. In the context of MAF, we utilize MCA2, which calculates context-aware multimodal key and value representations followed by traditional dot-product attention. 

Given an intermediate representation H generated by a pre-trained language model, query, key, and value vectors $Q$, $K$, and $V \in R^{n \times d}$ are calculated. Next, given a multmodal representation $C \in R^{c \times d_{c}}$, multimodal information informed key \& value vectors $\hat{K}$ \& $\hat{V}$ are obtained. Next, via a learnable matrix $\lambda \in R^{n \times 1}$ and project matrix $U_{k}, U_{v}$, we derive attentive key \& value representations with the amalgamation of both the modalities under consideration (See equation \ref{eq:maf_1}).

\begin{equation}
\begin{small}  
\begin{bmatrix} \hat{K} \\ \hat{V} \end{bmatrix} =
\left(1 - \begin{bmatrix} \lambda_k \\ \lambda_v \end{bmatrix}\right)
\begin{bmatrix} K \\ V \end{bmatrix} +
\begin{bmatrix} \lambda_k \\ \lambda_v \end{bmatrix} (C \begin{bmatrix} U_k \\ U_v \end{bmatrix})
\end{small}  
\label{eq:maf_1}
\end{equation}

Finally, the multimodal information-infused vectors $\hat{K}$ and $\hat{V}$ are used to compute the traditional scaled dot-product attention. Then, we define modality-specific gates, assigning weights to each modality-aware hidden representation concerning $H$, thus formulating the final information-fused representation $\hat{H}$ (refer to equation \ref{eq:maf_2}).

\begin{equation}
\hat{H}=H + g_{m_{1}} \odot H_{m_{1}} + \ldots
\label{eq:maf_2}
\end{equation}

Furthermore, as we use a decoder model to predict the next action sub-sequence, we propose employing a fusion technique that combines previously predicted actions with inferred camera observations through concatenation and linear transformation.

\end{document}